\documentclass[sigconf]{acmart}

\AtBeginDocument{%
  \providecommand\BibTeX{{%
    \normalfont B\kern-0.5em{\scshape i\kern-0.25em b}\kern-0.8em\TeX}}}

\begin{document}

\title{A Compare Aggregate Transformer for Understanding Document-grounded Dialogue}


\author{Longxuan Ma, Weinan Zhang, Runxin Sun, Ting Liu}\email{{lxma,wnzhang,rxsun,tliu}@ir.hit.edu.cn}
\affiliation{%
  \institution{Research Center for Social Computing and Information Retrieval, Harbin Institute of Technology}
  \city{Harbin}
  \state{Heilongjiang}
  \country{China}
}


\begin{abstract}
Unstructured documents serving as external knowledge of the dialogues help to generate more informative responses. Previous research focused on knowledge selection (KS) in the document with dialogue. However, dialogue history that is not related to the current dialogue may introduce noise in the KS processing. In this paper, we propose a Compare Aggregate Transformer (CAT) to jointly denoise the dialogue context and aggregate the document information for response generation. We designed two different comparison mechanisms to reduce noise (before and during decoding). In addition, we propose two metrics for evaluating document utilization efficiency based on word overlap. Experimental results on the CMUDoG dataset show that the proposed CAT model outperforms the state-of-the-art approach and strong baselines. 
\end{abstract}

\ccsdesc[500]{Computing methodologies~Natural language generation}
\ccsdesc[500]{Computing methodologies~Discourse, dialogue and pragmatics}

\keywords{Dialogue system, Natural language generation, knowledge selection}


\maketitle
\section{Introduction}
Dialogue system (DS) attracts great attention from industry and academia because of its wide application prospects. Sequence-to-sequence models (Seq2Seq) \cite{DBLP:conf/nips/SutskeverVL14,DBLP:conf/aaai/SerbanSBCP16} are verified to be an effective framework for the DS task. However, one problem of Seq2Seq models is that they tended to generate generic responses that provids deficient information \citet{DBLP:conf/naacl/LiGBGD16,DBLP:conf/aaai/GhazvininejadBC18}. Previous researchers proposed different methods to alleviate this issue. One way is to focus on models' ability to extract information from conversations. \citet{DBLP:conf/naacl/LiGBGD16} introduced Maximum Mutual Information (MMI) as the objective function for generating diverse response. \citet{DBLP:conf/aaai/SerbanSLCPCB17} proposed a latent variable model to capture posterior information of golden response. \citet{DBLP:conf/acl/ZhaoZE17} used conditional variational autoencoders to learn discourse-level diversity for neural dialogue models. The other way is introducing external knowledge, either unstructured knowledge texts \citet{DBLP:conf/aaai/GhazvininejadBC18,DBLP:journals/corr/abs-1903-09813,DBLP:conf/iclr/DinanRSFAW19} or structured knowledge triples \cite{DBLP:conf/acl/LiuFCRYL18,DBLP:conf/aaai/YoungCCZBH18,DBLP:conf/ijcai/ZhouYHZXZ18} to help open-domain conversation generation by producing responses conditioned on selected knowledge.


\begin{table}[t]
\small
\centering
\begin{tabular}{ll}  
\hline
Document:      \\
 Movie Name: The Shape of Water. ... Director: Guillermo del Toro. Gen-\\
 re: Fantasy, Drama.Cast: Sally Hawkins as Elisa Esposito, a mute cleaner\\
 who works at a secret government  laboratory. ... Critical Response: one\\
 of del Toro's most stunningly successful works ... \\
\hline
Dialogue: \\
\textbf{S1}: I thought The Shape of Water was one of Del Toro's best works. \\ 
What about you?\\
\textbf{S2}: \textit{\textcolor{red}{Yes, his style really extended the story}}.\\
\textbf{S1}: I agree. He has a way with fantasy elements that really helped this s-\\
tory be truly beautiful. It has a very high rating on rotten tomatoes, too. \\
\textbf{S2}: \textbf{Sally Hawkins acting was phenomenally expressive. Didn't fe-}\\
\textbf{el her character was mentally handicapped.}\\
\textbf{S1}: The characterization of her as such was ... off the mark. \\
\hline
\end{tabular}
\caption{One DGD example in the CMUDoG dataset. S1/S2 means Speaker-1/Speaker-2, respectively.}
\label{CMUDoG}
\end{table}


The Document-grounded Dialogue (DGD) \cite{DBLP:conf/emnlp/ZhouPB18,DBLP:conf/ijcai/ZhaoTWX0Y19,DBLP:conf/acl/LiNMFLZ19} is a new way to use external knowledge. It establishes a conversation mode in which relevant information can be obtained from the given document. The DGD systems can be used in scenarios such as talking over merchandise against the product manual, commenting on news reports, etc. One example of DGD is presented in Table \ref{CMUDoG}. Two interlocutors talk about the given document and freely reference the text segment during the conversation. 

To address this task, two main challenges need to be considered in a DGD model: 1) Determining which of the historical conversations are related to the current conversation, 2) Using current conversation and the related conversation history to select proper document information and to generate an informative response. Previous work \citet{DBLP:conf/naacl/AroraKR19,DBLP:conf/ijcai/ZhaoTWX0Y19,DBLP:conf/acl/QinGBLGDCG19,DBLP:journals/corr/abs-2005-06128,DBLP:journals/corr/abs-1908-09528} generally focused on selecting knowledge with all the conversations. However, the relationship between historical conversations and the current conversation has not been studied enough. For example, in Table \ref{CMUDoG}, the italics utterance from user1, "\textit{Yes, his style really extended the story.}", is related to dialogue history. While the black fold utterance from user1, "\textbf{Sally Hawkins acting was phenomenally expressive. Didn't feel her character was mentally handicapped.}", has no direct relationship with the historical utterances. when employing this sentence as the last utterance, the dialogue history is not conducive to generate a response. 

In this paper, we propose a novel Transformer-based \cite{DBLP:conf/nips/VaswaniSPUJGKP17} model for understanding the dialogues and generate informative responses in the DGD, named Compare Aggregate Transformer (CAT). Previous research \cite{DBLP:conf/acl/SankarSPCB19} has shown that the last utterance is the most important guidance for the response generation in the multi-turn setting. Hence we divide the dialogue into the last utterance and the dialogue history, then measure the effectiveness of the dialogue history. If the last utterance and the dialogue history are related, we need to consider all the conversations to filter the document information. Otherwise, the existence of dialogue history is equal to the introduction of noise, and its impact should be eliminated conditionally. For this purpose, on one side, the CAT filters the document information with the last utterance; on the other side, the CAT uses the last utterance to guide the dialogue history and employs the guiding result to filter the given document. We judge the importance of the dialogue history by comparing the two parts, then aggregate the filtered document information to generate the response. Experimental results show that our model can generate more relevant and informative responses than competitive baselines. When the dialogue history is less relevant to the last utterance, our model is verified to be even more effective. The main contributions of this paper are:

(1) We propose a compare aggregate method to determine the relationship between the historical dialogues and the last utterance. Experiments show that our method outperforms strong baselines on the CMUDoG dataset\footnote{The code and data will be released in Github.}.

(2) We propose two new metrics to evaluate the document knowledge utilization in the DGD. They are both based on N-gram overlap among generated response, the dialogue, and the document. 


\begin{figure*}[t]
\centering
\includegraphics[width=\linewidth]{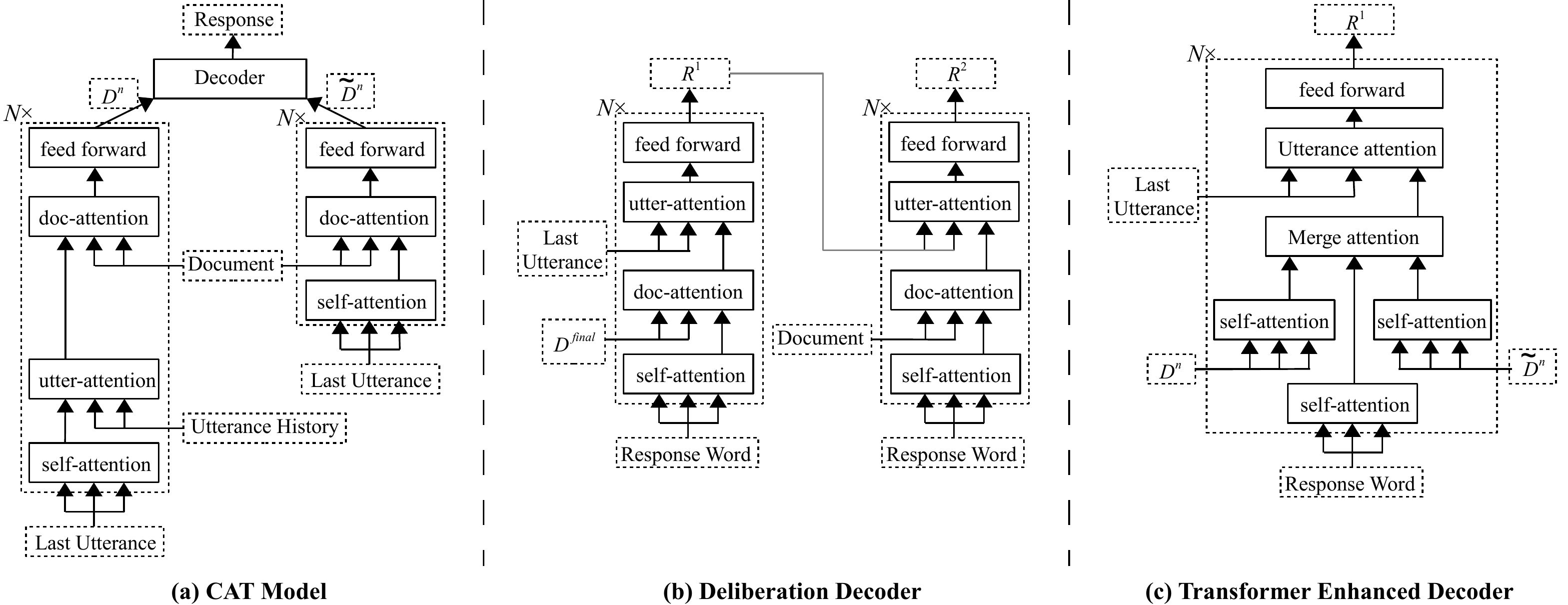}%
\caption{The architecture of the CAT model. "utter" is short for utterance. "doc" is short for document.}
\label{CAT}
\end{figure*}

\section{Related Work}

The DGD maintains a dialogue pattern where external knowledge can be obtained from the given document. Most recently, some DGD datasets \citet{DBLP:conf/emnlp/ZhouPB18,DBLP:conf/emnlp/MogheABK18,DBLP:conf/acl/QinGBLGDCG19,gopalakrishnan2019topical} have been released to exploiting unstructured document information in conversations. 


Models trying to address the DGD task can be classified into two categories based on their encoding process with dialogues: one is parallel modeling and the other is incremental modeling. For the first category, \citet{DBLP:conf/emnlp/MogheABK18} used a generation-based model that learns to copy information from the background knowledge and a span prediction model that predicts the appropriate response span in the background knowledge. \citet{DBLP:journals/corr/abs-1903-10245} claimed the first to unify knowledge triples and long texts as a graph. Then employed a reinforce learning process in the flexible multi-hop knowledge graph reasoning process. To improve the process of using background knowledge, \cite{DBLP:journals/corr/abs-1906-06685} firstly adopted the encoder state of the utterance history context as a query to select the most relevant knowledge, then employed a modified version of BiDAF \cite{DBLP:conf/iclr/SeoKFH17} to point out the most relevant token positions of the background sequence. 
\citet{DBLP:journals/corr/abs-1908-06449} used a decoding switcher to predict the probabilities of executing the reference decoding or generation decoding. Some other researchers \cite{DBLP:conf/ijcai/ZhaoTWX0Y19,DBLP:conf/naacl/AroraKR19,DBLP:conf/acl/QinGBLGDCG19,DBLP:journals/corr/abs-1908-06449,DBLP:journals/corr/abs-1908-09528} also followed this parallel encoding method. For the second category, \citet{DBLP:journals/corr/abs-2002-07510} proposed a sequential latent knowledge selection model for Knowledge-Grounded Dialogue. \citet{DBLP:conf/acl/LiNMFLZ19} designed an incremental transformer to encode multi-turn utterances along with knowledge in the related document. Meanwhile, a two-way deliberation decoder \cite{DBLP:conf/nips/XiaTWLQYL17} was used for response generation. However, the relationship between the dialogue history and the last utterance is not well studied. In this paper, we propose a compare aggregate method to investigate this problem. It should be pointed out that when the target response changes the topic, the task is to detect whether the topic is ended and to initiate a new topic \cite{DBLP:conf/naacl/AkasakiK19}. We do not study the conversation initiation problem in this paper, although we may take it as future work.

\section{The Proposed CAT Model}

\subsection{Problem Statement}
The inputs of the CAT model are the given document $\textbf{D}$ = ($D_1$, $D_2$, ..., $D_d$) with $d$ words, dialogue history $\textbf{H}$ = ($H_1$, $H_2$, ..., $H_h$) with $h$ words and the last utterance $\textbf{L}$ = ($L_1$, $L_2$, ..., $L_l$) with $l$ words. The task is to generate the response \textbf{R} = ($R_1$, $R_2$, ..., $R_r$) with $r$ tokens with probability:


\begin{align}
P(\textbf{R} | \textbf{H}, \textbf{L}, \textbf{D} ; \Theta ) =& \prod_{i=1}^r P(R_i | \textbf{H}, \textbf{L}, \textbf{D}, \textbf{R}_{<i}; \Theta),
\end{align}

where $\textbf{R}_{<i}$ = ($R_1$, $R_2$, ..., $R_{i-1}$), $\Theta$ is the model's parameters.

\subsection{Encoder}
The structure of the CAT model is shown in Figure \ref{CAT}. The hidden dimension of the CAT model is $\widehat{h}$. We use the Transformer structure \cite{DBLP:conf/nips/VaswaniSPUJGKP17}. The self-attention is calculated as follow:

\begin{align}
\text{Attention}(\textbf{Q}, \textbf{K}, \textbf{V}) =& \text{softmax}(\frac{\textbf{Q}  \textbf{K}^T}{\sqrt{d_k}})  \textbf{V},
\end{align}

where \textbf{Q}, \textbf{K}, and \textbf{V} are the query, the key, and the value, respectively; $d_k$ is the dimension of \textbf{Q} and \textbf{K}. The encoder and the decoder stack $N$ ($N=3$ in our work) identical layers of multihead attention (MAtt):

\begin{align}
\text{MAtt}(\textbf{Q}, \textbf{K}, \textbf{V}) =& [\textbf{A}_1, ... , \textbf{A}_n] \textbf{W}^O, \\
\textbf{A}_i = \text{Attention}(\textbf{Q}&  \textbf{W}^Q_i, \textbf{K} \textbf{W}^K_i, \textbf{V} \textbf{W}^V_i),
\end{align}

where $\textbf{W}^Q_i, \textbf{W}^K_i, \textbf{W}^V_i$ $(i = 1, ..., n )$ and $\textbf{W}^O$ are learnable parameters. 

The encoder of CAT consists of two branches as figure \ref{CAT} (a). The left branch learns the information selected by dialogue history \textbf{H}, the right part learns the information chosen by the last utterance \textbf{L}. After self-attention process, we get $\textbf{H}_{s} = \text{MAtt}(\textbf{H}, \textbf{H}, \textbf{H})$ and $\textbf{L}_{s} = \text{MAtt}(\textbf{L}, \textbf{L}, \textbf{L})$. Then we employ $\textbf{L}_{s}$ to guide the \textbf{H}. $\textbf{H}^{1} = \text{MAtt}(\textbf{L}_{s}, \textbf{H}, \textbf{H})$, where $\textbf{H}^{1}$ is the hidden state at the first layer. Then we adopt $\textbf{H}^{1}$ to select knowledge from the document \textbf{D}, $\textbf{D}^{1} = \text{FF}(\text{MAtt}(\textbf{H}^{1}, \textbf{D}, \textbf{D}))$. FF is the feed-forward process. In the second layer, $\textbf{D}^{1}$ is the input, $\textbf{D}_{s}^{1} = \text{MAtt}(\textbf{D}^{1}, \textbf{D}^{1}, \textbf{D}^{1}))$, $\textbf{H}^{2} = \text{MAtt}(\textbf{D}_{s}^{1}, \textbf{H}, \textbf{H})$, $\textbf{D}^{2} = \text{FF}(\text{MAtt}(\textbf{H}^{2}, \textbf{D}, \textbf{D}))$. After $N$ layers, we obtain the information $\textbf{D}^{n}$ selected by \textbf{H}. In the right branch, we use $\textbf{L}_{s}$ to filter the \textbf{D}. $\widetilde{\textbf{D}^{n}}$ is the information selected by \textbf{L}. 





\subsection{Comparison Aggregate}
As demonstrated by \cite{DBLP:conf/acl/SankarSPCB19}, the last utterance played an fundamental role in response generation. We need to preserve the document information filtered by \textbf{L}, and determine how much information selected by \textbf{H} is needed. We propose $2$ different compare aggregate methods: one is concatenation before decoding and the other is attended comparison in the decoder.

\subsubsection{Concatenation}
We use average pooling to $\textbf{H}_s$ and $\textbf{L}_s$ to get their vector representations $\textbf{H}_{sa}$ and $\textbf{L}_{sa} \in \mathbb{R}^{\widehat{h}*1}$, respectively. The concatenation method calculates relevance score $\alpha$ to determine the importance of $\textbf{D}^{n}$ as follow:

\begin{align}
\alpha =&\text{tanh}{(\textbf{H}_{sa} \textbf{W}^{H} + \textbf{L}_{sa} \textbf{W}^{L})}, \\
\textbf{D}_{final} =& [ \text{sigmoid}( \textbf{W}^{\alpha} \alpha ) * \textbf{D}^{n}; \widetilde{ \textbf{D}}^{n}  ],
\end{align}

where $\textbf{W}^{H}$, $\textbf{W}^{L} \in \mathbb{R}^{\widehat{h}*\widehat{h}}$, $\textbf{W}^{\alpha} \in \mathbb{R}^{1*\widehat{h}}$ are learnable parameters. $[\textbf{X};\textbf{Y}]$ is the concatenation of \textbf{X} and \textbf{Y} in sentence dimension. $*$ is the element-wise multiplication. Note that the $\textbf{D}^{n}$ is guided by \textbf{H}, the concatenation method performs a second level comparison with $\textbf{H}$ and $\textbf{L}$ and then transfers the topic-aware $\textbf{D}_{final}$ to the two-pass Deliberation Decoder (DD) \cite{DBLP:conf/nips/XiaTWLQYL17}. The structure of the DD is shown in Figure \ref{CAT} (b). The first-pass takes \textbf{L} and $\textbf{D}_{final}$ as inputs and learns to generate a contextual coherently response $\textbf{R}^1$. The second-pass takes $\textbf{R}^1$ and the document \textbf{D} as inputs and learns to inject document knowledge. The DD aggregates document, conversation, and topic information to generate the final response $\textbf{R}^2$. Loss is from both the first and the second layers:


\begin{align}
L =&  - \sum_{m=1}^{M} \sum_{i=1}^{r} ( \text{log}P(R^1_i) + \text{log}P(R^2_i) ),
\end{align}

where $M$ is the total training example; $R^1_i$ and $R^2_i$ are the $i$-th word generated by the first and second decoder layer, respectively. 


\subsubsection{Attended Comparison}
We employ an Enhanced Decoder \cite{DBLP:conf/cikm/ZhengZ19} to perform the attended comparing. The structure of our Enhanced Decoder is illustrated in Figure \ref{CAT} (c). It accepts $\textbf{D}^{n}$, $\widetilde{ \textbf{D}}^{n} $ and the response \textbf{R} as inputs, applying a different way to compare and aggregate. The merge attention computes weight across all inputs:

\begin{align}
\textbf{P} = & [ \textbf{R}; \textbf{D}^{n}; \widetilde{ \textbf{D}}^{n}  ] \textbf{W}_P,\\
\textbf{V}_{merge} = & P_R  \textbf{R} + P_D  \textbf{D}^{n} + P_{\widetilde{ D}}  \widetilde{ \textbf{D}}^{n} ,
\end{align}

where $W_P$ is learnable parameters. The dimension of $P$ is $3$. $P_R$, $P_D$ and $P_{\widetilde{D}}$ are the Softmax results of \textbf{P}. $\textbf{V}_{merge}$ and \textbf{L} are used for next utterance attention as shown in Figure \ref{CAT} (c). The output of the Enhanced Decoder is connected to the second layer of DD and we define this new structure as Enhanced Deliberation Decoder (EDD). The loss is the same as Eq. (7). 

\section{Experiments}

\subsection{Dataset}
We evaluate our model with the CMUDoG \cite{DBLP:conf/emnlp/ZhouPB18} dataset. There are $4112$ dialogs based on $120$ documents in the dataset. One document contains $4$ sections, such as movie introduction and scenes. A related section is given for every several consequent utterances. However, the conversations are not constrained to the given section. In our setting, we use the full document (with 4 section) as external knowledge. The average length of documents is around $800$ words. We concatenate consequent utterances of the same person as one utterance. When training, we remove the first two or three rounds of greeting sentences. Each sample contains one document, two or more historical utterances, one last utterance, and one golden response. When testing, we use two different versions of the test set. The first follows the process of training data, we name it Reduced version. The second is constructed by comparing the original document section of the conversation based, we preserve the examples that the dialogue history and the last utterance are based on different document sections. For example, dialogue history is based on section $2$, the last utterance and response are based on section $3$. We name it Sampled version and it is used for testing our models' comprehending ability of the topic transfer in conversations. The data statistics are shown in Table \ref{statistics}. Please refer to \citet{DBLP:conf/emnlp/ZhouPB18} for more details. It is worth noting that the sampled version does not represent the proportion of all conversation topic transfers, but it demonstrates this problem better than the Reduced version. 

\begin{table}[t]
\centering
\begin{tabular}{l|c|c}  
\hline
Dataset   & Utterances (train / dev / test)  & Word/Utterance \\
\hline
Original   & 72922 / 3626 / 11577  & 18.6 \\
\hline
Reduced  & 66332 / 3269 / 10502  & 19.7 \\
\hline
Sampled  & 66332 / 3269 / 1317 & 19.6 \\
\hline
\end{tabular}
\caption{Statistics of the CMUDoG dataset.}
\label{statistics}
\end{table}


\subsection{Baselines}
We evaluated several competitive baselines.

\subsubsection{RNN-based models}



VHRED: A Hierarchical Latent Variable Encoder-Decoder Model \cite{DBLP:conf/aaai/SerbanSLCPCB17}, which introduces a global (semantic level) latent variable $Z$ for the problem that HRED \cite{DBLP:conf/aaai/SerbanSBCP16} is difficult to generate meaningful and high-quality replies. $Z$ is calculated with the encoder RNN outputs and the context RNN outputs. The latent variable $Z$ contains some high-level semantic information, which encourages the model to extract abstract semantic concepts. Please refer to \citet{DBLP:conf/aaai/SerbanSLCPCB17} for more details. We use $Z$ to capture the topic transfer in conversations and test three different settings. For the first setting, we do not employ the document knowledge, only use dialogue as input to generate the response. It is recorded as VHRED(-k). For the second one, we use the same encoder RNN with shared parameters to learn the representation of the document and the utterance, then concatenate the final hidden state of them as the input of the context RNN. It is denoted by VHRED(c). For the third one, we use word-level dot-attention \cite{DBLP:conf/emnlp/LuongPM15} to get the document-aware utterance representation and use it as the input of context RNN. It is termed as VHRED(a).



\subsubsection{Transformer-based models}

T-DD/T-EDD: They both use the Transformer as the encoder. The inputs are the concatenation of dialogues and the document. These two models parallel encode the dialogue without detecting topic transfer. The T-DD uses a Deliberation Decoder (DD) as the decoder. The T-EDD uses an Enhanced Deliberation Decoder (EDD) as the decoder. 

ITDD \cite{DBLP:conf/acl/LiNMFLZ19}: It uses Incremental Transformer Encoder (ITE) and two-pass Deliberation Decoder (DD). Incremental Transformer uses multi-head attention to incorporate document sections and context into each utterance’s encoding process. ITDD incrementally models dialogues without detecting topic transitions. 


\begin{table*}[t]
\small
%
\begin{center} 
\begin{tabular}{l|c|c|c|c|c}
\hline
Model   & PPL    &BLEU (\%)     & ROUGE-L    & KU-2/3 (\%) & QKU-2/3 \\
\hline
VHRED(-k) & 97.3$\diamond$ (99.3)* &   0.49* (0.49)*   & 7.80* (7.82)*   & --/--  (--/--) & --/--  (--/--) \\

VHRED(c) & 80.2$\diamond$ (85.4)* &   0.79* (0.77)*    & 8.64* (8.63)* & 12.0/27.0$\diamond$ (12.1/27.6)$\diamond$  &  3.36/2.82$\diamond$ (3.35/2.80)$\diamond$ \\

VHRED(a) & 77.2$\diamond$ (78.5)*&    0.84* (0.80)*    & 8.98* (8.99)* & \textbf{13.7/31.7}$\diamond$ (\textbf{13.1/31.3})*  &   3.23/2.72* (3.23/2.72)* \\
\hline

T-DD   & 18.2* (20.5)* &   0.90* (0.89)*   & 9.23* (9.24)*& 8.0/23.1* (8.0/23.0)*  &  2.55/1.94* (2.55/1.95)* \\
T-EDD  & 18.2* (20.3)* &   0.91* (0.90)*   & 9.35* (9.36)* & 8.3/23.5* (8.1/23.4)* & 2.45/1.91* (2.45/1.92)* \\
ITDD & 16.2* (18.7)* &   1.01* (0.99)*   & 10.12$\diamond$ (10.10)*  &  9.0/24.5* (9.1/24.4)* &  2.18/1.84* (2.15/1.82)* \\
\hline

CAT-EDD & 16.0* (18.2)*&    1.14* (1.14)*   & 11.10* (11.12)*& 9.5/24.8* (9.7/24.9)* &  2.12/1.77* (2.11/1.76)*  \\
CAT-DD  & \textbf{15.2} (\textbf{16.1}) &   \textbf{1.22 (1.21)}   & \textbf{11.22} (\textbf{11.22}) & 11.0/26.5 (11.1/26.4) &  \textbf{2.08/1.64} (\textbf{2.05/1.62})  \\
\hline
\end{tabular}
\caption{Automatic evaluations on the CMUDoG Dataset. $\cdot$ ($\cdot$) means Reduced (Sampled) test data. We take the CAT-DD as the base model to do the significant test, $\diamond$ and * stands p$<$0.05 and p$<$0.01, respectively. }
\label{auto-random}
\end{center} 
\end{table*}

\subsection{Evaluation Metrics}

\textbf{Automatic Evaluation}: We employ perplexity (PPL) \cite{DBLP:conf/nips/BengioDV00}, BLEU \cite{DBLP:conf/acl/PapineniRWZ02} and ROUGE \cite{lin2004rouge}. The PPL of the gold response is measured, lower perplexity indicates better performance. BLEU measures the n-gram overlap between a generated response and a gold response. Since there is only one reference for each response, BLEU scores are extremely low. ROUGE measures the n-gram overlap based on the recall rate. Since the conversations are constrained by the background material, ROUGE is reliable. 

We also introduce two metrics to automatically evaluate the \textbf{Knowledge Utilization (KU)}, they are both based on $N$-grams overlaps. We define one document, conversations and generated response in Test set as (\textbf{D},\textbf{C},\textbf{R}). The $N$-grams set of each (\textbf{D},\textbf{C},\textbf{R}) are termed as $\textbf{G}_d^{N}, \textbf{G}_c^{N}$ and $\textbf{G}_r^{N}$, respectively. The number of overlapped $N$-grams of $\textbf{G}_d^{N}$ and $\textbf{G}_r^{N}$ is recorded as $\textbf{G}_{dr}^{N}$. Tuples which are in $\textbf{G}_{dr}^{N}$ but not in $\textbf{G}_c^{N}$ is named $\textbf{G}_{dr-c}^{N}$. Then $\textbf{KU} = len(\textbf{G}_{dr-c}^{N})/len(\textbf{G}_{dr}^{N}) $ reflects how many $N$-grams in the document are used in the generated replies, $len(\textbf{G})$ is the tuple number in \textbf{G}. The larger the KU is, the more $N$-grams of the document is utilized. Since low-frequency tuples may be more representative of text features, we define the reciprocal of the frequency of each tuple $k$ in \textbf{G} as $\textbf{R}_k^{G}$, which represents the importance of a tuple. Then the \textbf{Quality of Knowledge Utilization (QKU)} is calculated as:

\begin{align}
\textbf{QKU} = \sum_{(\textbf{D},\textbf{C},\textbf{R})} \frac{\sum_k{\textbf{R}_k^{G_r}}}{\sum_k{\textbf{R}_k^{G_d}}}, \quad k \in \textbf{G}_{dr-c}.
\end{align}

If $\textbf{R}_{k}^{G_r}$ is more important in response and $\textbf{R}_{k}^{G_d}$ is less important in document, the QKU will become even larger. So the smaller QKU means the higher quality of the used document knowledge. 

\textbf{Human Evaluation}: We randomly sampled $100$ conversations from the Sampled test set and obtained $800$ responses from eight models. We have 5 graduate students as judges. They score each response with access to previous dialogues and the document. We use three metrics: Fluency, Coherence, and Informativeness. Fluency measures whether the response is a human-like utterance. Coherence measures if the response is coherent with the dialogue context. Informativeness measures if the response contains relevant and correct information from the document. They are scored from 1 to 5 (1:very bad, 2:bad, 3:acceptable, 4:good, 5:very good). Overall inter-rater agreement measured by Fliess’ Kappa is $0.32$ ("fair"). 


\subsection{Experimental Setup} 
We use OpenNMT-py \cite{klein-etal-2017-opennmt} as the code framework. For all models, the pre-trained $300$ dimension word embedding \cite{DBLP:conf/nips/MikolovSCCD13} is shared by dialogue, document, and generated responses, the dimension of the hidden size is $300$. For the RNN-based models, $3$-layer bidirectional GRU and $3$-layer GRU are applied for encoder and decoder, respectively. For the Transformer-based models, the layers of both encoder and decoder are set to $3$, the number of heads in multi-head attention is $8$ and the filter size is $2048$. We use Adam ($\alpha$ = 0.001, $\beta_1$ = 0.9, $\beta_2$ = 0.999, and $\epsilon$ = $10^{-8}$) \cite{DBLP:journals/corr/KingmaB14} for optimization. The beam size is set to $5$ in the decoder. We truncate the words of the document to $800$ and the dialogue utterance to $40$. All models are trained on a TITAN X (Pascal) GPU. The average training time per epoch is around $40$ minutes for the Transformer-based models and around $20$ minutes for the RNN-based models. 

\section{Analysis}

\subsection{Experimental Results study}
Table \ref{auto-random} shows the automatic evaluations for all models on the Reduced (Sampled) dataset. The dialogue history is $2$ rounds. We only present ROUGE-L as ROUGE-1/2 show the same trend as ROUGE-L. Through experiments, we can see that the change range of KU-2 ($8.0$-$13.7$) is less than KU-3 ($23.1$-$31.7$) on the Reduced data, indicating that the KU-3 can better reflect the amount of knowledge used than KU-2. 

In the RNN-based models, the VHRED(-k) gets the worst PPL/ BLEU/ ROUGE, which reveals the importance of injecting document knowledge in the DGD task. We did not calculate the KU/QKU of the VHRED(-k) since the model did not use document knowledge. The VHRED(a) gets better PPL/BLEU/ROUGE/KU/QKU than the VHRED(c) model, which means the smaller granular extraction of document information benefits more in generating responses. 

Among the Transformer-based models, The ITDD model gets better PPL/BLEU/ROUGE-L/KU/QKU than the T-DD model, which means the incremental encoding method is stronger than parallel encoding. The CAT-EDD and the CAT-DD models achieve better performance than the T-DD and the T-EDD models, respectively. It indicates that our Compare-Aggregate method is helpful to understand the dialogue. The CAT-EDD model outperforms the ITDD model on all metrics, which indicates that our CAT module automatically learns the topic transfer between conversation history and the last utterance as we expected. The CAT-EDD does not perform as good as the CAT-DD, which shows that it is necessary to set up an independent mechanism to learn topic transfer, rather than automatic learning by attentions in the decoder. 

Comparing with the RNN-based models, the Transformer-based models get better performance on PPL/BLEU/ROUGE. It proves that the latter is better in the ability of convergence to the ground truth. The VHRED(c) and the VHRED(a) get better KU and worse QKU than the Transformer-based models. It means that the latent variable models increase the diversity of replies and use more document tuples, but their ability to extract unique tuples is not as good as the Transformer-based ones.



\begin{table}[t]
\small
\begin{center} 
\begin{tabular}{l|c|c|c}
\hline
Model & Flu. & Coh. &   Inf.  \\
\hline
VHRED(-k)    & 3.71 (3.72) & 2.82 (2.72) & 3.01 (2.82)\\

VHRED(c)     & 3.73 (3.82) & 3.04 (3.11) & 3.03 (3.05)\\

VHRED(a)    & 3.84 (3.77) & 3.11 (3.14) & 3.22 (3.06)\\
\hline
T-DD     & 3.84 (3.82) & 3.03 (3.06) & 3.03 (3.06)\\ 
T-EDD    & 3.84 (3.83) & 3.02 (3.08) & 3.05 (3.05)\\
ITEDD    & 3.90 (3.91) & 3.11 (3.12) & 3.43 (3.42)\\ 
\hline
CAT-EDD  & 4.02 (3.93) & 3.12 (3.33) & 3.33 (3.41)  \\
CAT-DD  & \textbf{4.09} (\textbf{4.09}) & \textbf{3.39} (\textbf{3.43}) & \textbf{3.44} (\textbf{3.61}) \\

\hline
\end{tabular}
\caption{Manual evaluations on the CMUDoG Dataset. Flu. /Coh. /Inf. /$\cdot$ ($\cdot$) mean Fluency /Coherence /Informativeness /Reduced (Sampled) test data, respectively.}
\label{manual-random}
\end{center} 
\end{table}

Table \ref{manual-random} shows the manual evaluations for all models on the Reduced(Sampled) dataset. The CAT-DD model gets the highest scores on Fluency/Coherence/Informativeness. When experimenting with the Sampled test set, we can see that the advantages of our models become greater than the results of the Reduced version in both automatic and manual evaluations. Our model shows more advantages in datasets with more topic transfer. 

\begin{table}[t]
\centering
\begin{tabular}{l|c|c|c}
\hline
Models & PPL & BLEU & KU-2(\%)/QKU-2\\
\hline
CAT-DD & 16.1 &  1.21 & 11.1 / 2.05\\
\hline
w/o-left & 19.8* & 0.90* & 8.2* / 2.56* \\
w/o-(5,6) & 18.7* & 0.93* & 9.1* / 2.48$\diamond$ \\
w/o-(G)& 18.2* & 0.96* &9.2$\diamond$ / 2.46* \\
\hline
\end{tabular}
\caption{Ablation Study on the Sampled test set. We take the CAT-DD as the base model to do the significant test, $\diamond$ and * stand for p$<$0.05 and p$<$0.01, respectively. w/o = without.}
\label{ablation-random}
\end{table}

\subsection{Ablation Study}
Table \ref{ablation-random} illustrates the ablation study of the CAT-DD model. w/o-left means the left branch is removed and the model degenerates to T-DD which takes the last utterance and document as inputs. We can see that all the automatic evaluation indexes significantly reduce, indicating the dialogue history can not be simply ignored. w/o-(5,6) is a model without Eq. (5) and (6), which is equivalent to simply connect the outputs of the left and the right encoder branches. The results showed that the ability of the model to distinguish the conversation topic transfer is weakened. w/o-(G) is a model removing the utter-attention in the left branch, which means we \textbf{do not use L to guide the H}, the structure of left branch changes to the right branch and the input is \textbf{H}. The performance is declining, which indicates that the guiding process is useful. The significant tests (two-tailed student t-test) on PPL/BLEU/KU-2/QKU-2 reveal the effectiveness of each component. 

\begin{figure}[t]
\centering
\includegraphics[width=2.8in]{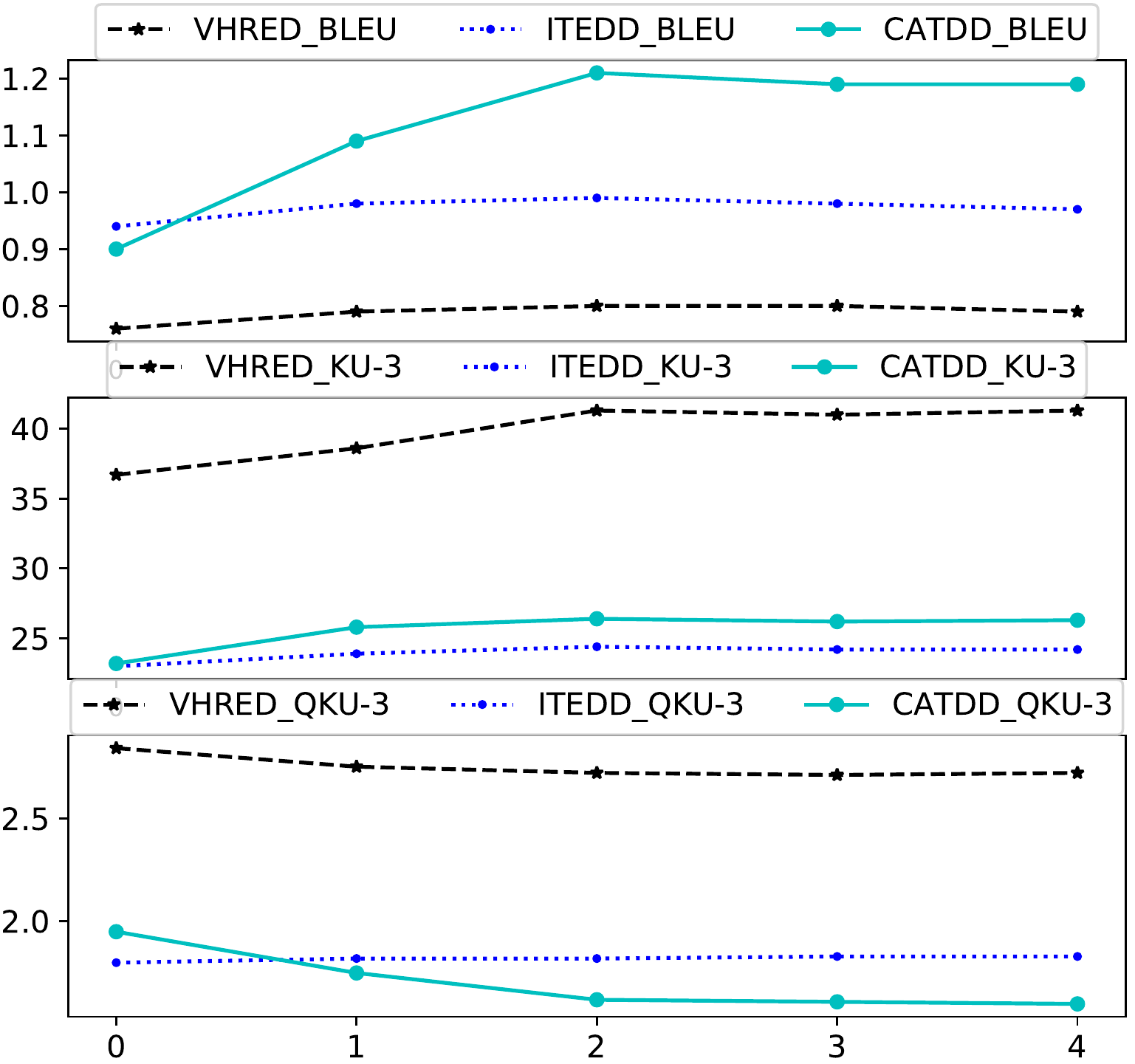}
\caption{The effect of dialogue history rounds on VHRED(a)/ITDD/CAT-DD models. The abscissa represents the historical dialogue rounds. The ordinate represents the BLEU/KU-3/QKU-3 values.}
\label{CAT-round}
\end{figure}

\subsection{History Round Study}

We use the CAT-DD model and the Sampled test set to study the influence of the historical dialogue rounds. For example, setting dialogue history to $0$ means we use only the last utterance, the CAT-DD becomes the w/o-left model in the ablation study. Setting dialogue history to $N$ means we use $N$ rounds of dialogue history for the input of the left branch. We set the conversation history to $0/1/2/3/4$ to test the response of VHRED(a)/ITDD/CAT-DD models. Figure \ref{CAT-round} shows the trend of BLEU/KU-3/QKU-3. The top figure shows the BLEU trend, the CAT-DD reaches the maximum when the rounds are $2$. The continuous increase of rounds does not significantly improve the generation effect. In the middle picture, with the increase of historical dialogue from $0$ to $2$, the VHRED(a) and the CAT-DD have a visible improvement on the KU-3, which shows that the information contained in the historical dialogue can be identified and affect the extraction of document information. The ITDD model is not as sensitive as the others on the KU-3, indicating that the incremental encoding structure pays more attention to the information of the last utterance. The bottom figure shows the trend of the QKU-3. When the history dialogue increases, the ITDD model keeps stable and the VHRED(a) and the CAT-DD models have a declining trend, which again indicates that the VHRED(a) and the CAT-DD are more sensitive to the historical dialogue. 



\begin{figure}[t]
\centering
\includegraphics[width=2.8in]{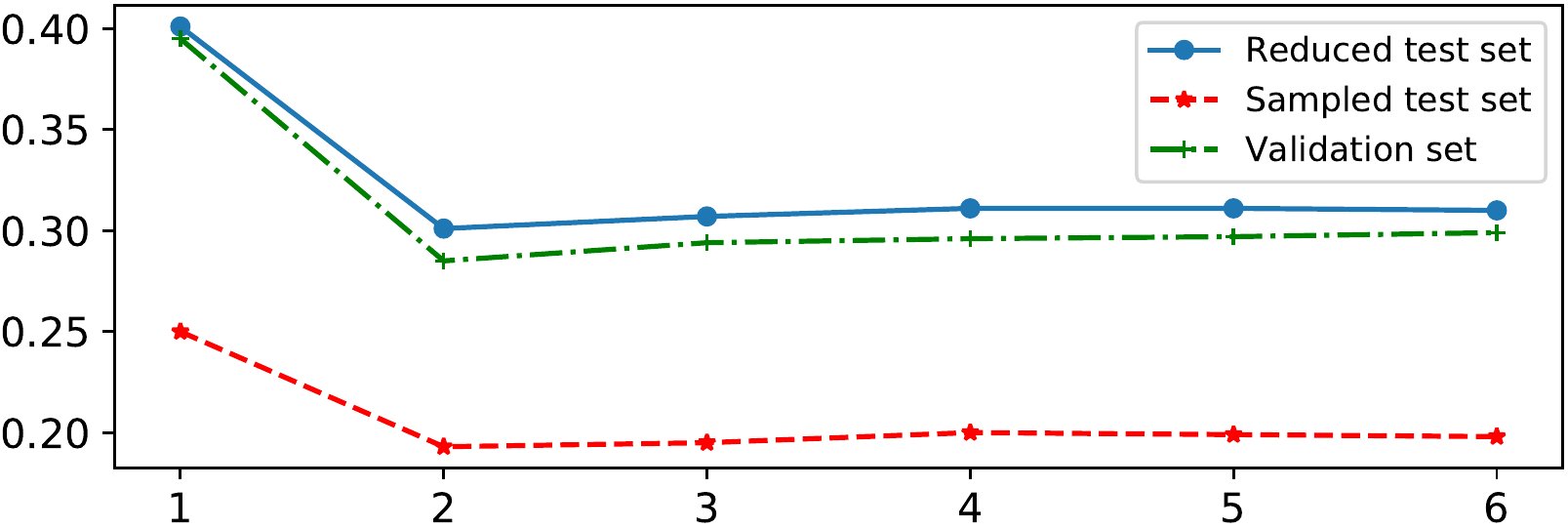}
\caption{The rating of dialogue history in the CAT-DD model with Reduced and Sampled test set. The abscissa represents the dialogue rounds and the ordinate represents the correlation score in the model.}
\label{CAT-alpha}
\end{figure}

\subsection{History Importance Study}
Figure \ref{CAT-alpha} shows the average sigmoid$( W^{\alpha} \alpha )$ value in the CAT-DD model over the Reduced/Sampled test set and the Validation set. A higher value means a stronger correlation between the last utterance and the historical dialogue. We can see that on the Reduced test set and the Validation set, the relevance score is higher than that of the Sampled data, which proves that the last utterance and the historical dialogue are more irrelevant in the latter. Our model captures this change and performs better on the Sampled data than the Reduced data. When the historical rounds increase from $1$ to $2$, the relevance score reduces obviously for all data sets, which means the increase of dialogue history introduces more unrelated information. When the historical conversations increases from $2$ to $6$, all data have no significant change,  indicating that increasing the dialogue rounds does not improve the recognition ability of the model to the topic change. 

\begin{table}[t]
\small
\begin{center} 
\begin{tabular}{l}
\hline
Document:\\
... sally hawkins as elisa esposito, a mute cleaner who works at\\
a secret government laboratory. michael shannon as colonel\\
richard strickland ... rating rotten tomatoes: 92\% The shape of\\
water is a 2017 american fantasy film ... it stars \textcolor{red}{sally hawkins,}\\
\textcolor{red}{michael shannon, richard jenkins, Doug jones, michael stuhl-}\\, \textcolor{red}{barg and octavia spencer} ... \\
\hline
Dialogue history:\\
\textbf{S1}: I wonder if it's a government creation or something\\
captured from the wild. i would assume the wild. \\
\textbf{S2}: It was captured for governmental experiments.\\
\hline
The last Utterance: \textbf{S1}: \textcolor{blue}{Is it a big name cast?}\\
\hline
Groud truth:\\
\textbf{S2}: Sally hawkins played the role of the mute cleaner. while \\
michael shannon played the role of colonel richard strickland.\\
\hline
Generated response:\\
\textbf{VHRED(a)}: it has rating rotten tomatoes: 92\% \\
\textbf{TDD}: i am not sure about it. \\
\textbf{ITDD}: yes, sally hawkins as elisa esposito. \\
\textbf{CAT-DD}: \textcolor{red}{sally hawkins, michael shannon, richard jenkins,}\\
 \textcolor{red}{doug jones, michael stuhlbarg, and octavia spencer.} \\
\textbf{(w/o-(5,6))}: sally hawkins works at a government laboratory.\\
\textbf{(w/o-(G))}: it is a 2017 american fantasy film.\\
\hline
\end{tabular}
\caption{Case study in the CMUDoG Sampled Dataset. S1/S2 means Speaker-1/Speaker-2, respectively. (w/o-(5,6)) and (w/o-(G)) are models in the ablation study.}
\label{examples-random}
\end{center} 
\end{table}

\subsection{Case Study}
In Table \ref{examples-random}, we randomly select an example in the Sampled test set for a case study. The document, the dialogue history, the last utterance, and the ground truth are presented. We can observe that the last utterance is irrelevant to the dialogue history. The generated responses of different models are listed below. The VHRED(a) and CAT-DD(w/o-(G)) models misunderstand the dialogue and use the wrong document knowledge. The TDD gives a generic reply. The ITDD model answers correctly but without enough document information. The CAT-DD(w/o-(5,6)) model gives a response that was influenced by the irrelevant historical dialogue which we want to eliminate. Only the CAT-DD model generates a reasonable reply and uses the correct document knowledge, which means it correctly understands the dialogues. 

\section{Conclusion}%

We propose the Compare Aggregate method to understand Document-grounded Dialogue (DGD). The dialogue is divided into the last utterance and the dialogue history. The relationship between the two parts is analyzed to denoise the dialogue context and aggregate the document information for response generation. Experiments show that our model outperforms previous work in both automatic and manual evaluations. Our model can better understand the dialogue context and select proper document information for response generation. We also propose Knowledge Utilization (KU) and Quality of Knowledge Utilization (QKU), which are used to measure the quantity and quality of the imported external knowledge, respectively. In the future, we will further study the topic transition problem and the knowledge injecting problem in the DGD. 


\bibliographystyle{ACM-Reference-Format}
\bibliography{sample-base}


\begin{thebibliography}{37}


\ifx \showCODEN    \undefined \def \showCODEN     #1{\unskip}     \fi
\ifx \showDOI      \undefined \def \showDOI       #1{#1}\fi
\ifx \showISBNx    \undefined \def \showISBNx     #1{\unskip}     \fi
\ifx \showISBNxiii \undefined \def \showISBNxiii  #1{\unskip}     \fi
\ifx \showISSN     \undefined \def \showISSN      #1{\unskip}     \fi
\ifx \showLCCN     \undefined \def \showLCCN      #1{\unskip}     \fi
\ifx \shownote     \undefined \def \shownote      #1{#1}          \fi
\ifx \showarticletitle \undefined \def \showarticletitle #1{#1}   \fi
\ifx \showURL      \undefined \def \showURL       {\relax}        \fi
\providecommand\bibfield[2]{#2}
\providecommand\bibinfo[2]{#2}
\providecommand\natexlab[1]{#1}
\providecommand\showeprint[2][]{arXiv:#2}

\bibitem[\protect\citeauthoryear{Akasaki and Kaji}{Akasaki and Kaji}{2019}]%
        {DBLP:conf/naacl/AkasakiK19}
\bibfield{author}{\bibinfo{person}{Satoshi Akasaki} {and}
  \bibinfo{person}{Nobuhiro Kaji}.} \bibinfo{year}{2019}\natexlab{}.
\newblock \showarticletitle{Conversation Initiation by Diverse News Contents
  Introduction}. In \bibinfo{booktitle}{\emph{{NAACL-HLT} {(1)}}}.
  \bibinfo{publisher}{Association for Computational Linguistics},
  \bibinfo{pages}{3988--3998}.
\newblock


\bibitem[\protect\citeauthoryear{Arora, Khapra, and Ramaswamy}{Arora
  et~al\mbox{.}}{2019}]%
        {DBLP:conf/naacl/AroraKR19}
\bibfield{author}{\bibinfo{person}{Siddhartha Arora},
  \bibinfo{person}{Mitesh~M. Khapra}, {and} \bibinfo{person}{Harish~G.
  Ramaswamy}.} \bibinfo{year}{2019}\natexlab{}.
\newblock \showarticletitle{On Knowledge distillation from complex networks for
  response prediction}. In \bibinfo{booktitle}{\emph{{NAACL-HLT} {(1)}}}.
  \bibinfo{publisher}{Association for Computational Linguistics},
  \bibinfo{pages}{3813--3822}.
\newblock


\bibitem[\protect\citeauthoryear{Bengio, Ducharme, and Vincent}{Bengio
  et~al\mbox{.}}{2000}]%
        {DBLP:conf/nips/BengioDV00}
\bibfield{author}{\bibinfo{person}{Yoshua Bengio},
  \bibinfo{person}{R{\'{e}}jean Ducharme}, {and} \bibinfo{person}{Pascal
  Vincent}.} \bibinfo{year}{2000}\natexlab{}.
\newblock \showarticletitle{A Neural Probabilistic Language Model}. In
  \bibinfo{booktitle}{\emph{{NIPS}}}. \bibinfo{publisher}{{MIT} Press},
  \bibinfo{pages}{932--938}.
\newblock


\bibitem[\protect\citeauthoryear{Dinan, Roller, Shuster, Fan, Auli, and
  Weston}{Dinan et~al\mbox{.}}{2019}]%
        {DBLP:conf/iclr/DinanRSFAW19}
\bibfield{author}{\bibinfo{person}{Emily Dinan}, \bibinfo{person}{Stephen
  Roller}, \bibinfo{person}{Kurt Shuster}, \bibinfo{person}{Angela Fan},
  \bibinfo{person}{Michael Auli}, {and} \bibinfo{person}{Jason Weston}.}
  \bibinfo{year}{2019}\natexlab{}.
\newblock \showarticletitle{Wizard of Wikipedia: Knowledge-Powered
  Conversational Agents}. In \bibinfo{booktitle}{\emph{{ICLR} (Poster)}}.
  \bibinfo{publisher}{OpenReview.net}.
\newblock


\bibitem[\protect\citeauthoryear{Ghazvininejad, Brockett, Chang, Dolan, Gao,
  Yih, and Galley}{Ghazvininejad et~al\mbox{.}}{2018}]%
        {DBLP:conf/aaai/GhazvininejadBC18}
\bibfield{author}{\bibinfo{person}{Marjan Ghazvininejad},
  \bibinfo{person}{Chris Brockett}, \bibinfo{person}{Ming{-}Wei Chang},
  \bibinfo{person}{Bill Dolan}, \bibinfo{person}{Jianfeng Gao},
  \bibinfo{person}{Wen{-}tau Yih}, {and} \bibinfo{person}{Michel Galley}.}
  \bibinfo{year}{2018}\natexlab{}.
\newblock \showarticletitle{A Knowledge-Grounded Neural Conversation Model}. In
  \bibinfo{booktitle}{\emph{{AAAI}}}. \bibinfo{publisher}{{AAAI} Press},
  \bibinfo{pages}{5110--5117}.
\newblock


\bibitem[\protect\citeauthoryear{Gopalakrishnan, Hedayatnia, Chen, Gottardi,
  Kwatra, Venkatesh, Gabriel, Hakkani-T{\"u}r, and AI}{Gopalakrishnan
  et~al\mbox{.}}{2019}]%
        {gopalakrishnan2019topical}
\bibfield{author}{\bibinfo{person}{Karthik Gopalakrishnan},
  \bibinfo{person}{Behnam Hedayatnia}, \bibinfo{person}{Qinlang Chen},
  \bibinfo{person}{Anna Gottardi}, \bibinfo{person}{Sanjeev Kwatra},
  \bibinfo{person}{Anu Venkatesh}, \bibinfo{person}{Raefer Gabriel},
  \bibinfo{person}{Dilek Hakkani-T{\"u}r}, {and} \bibinfo{person}{Amazon~Alexa
  AI}.} \bibinfo{year}{2019}\natexlab{}.
\newblock \showarticletitle{Topical-Chat: Towards Knowledge-Grounded
  Open-Domain Conversations}.
\newblock \bibinfo{journal}{\emph{Proc. Interspeech 2019}}
  (\bibinfo{year}{2019}), \bibinfo{pages}{1891--1895}.
\newblock


\bibitem[\protect\citeauthoryear{Kim, Ahn, and Kim}{Kim et~al\mbox{.}}{2020}]%
        {DBLP:journals/corr/abs-2002-07510}
\bibfield{author}{\bibinfo{person}{Byeongchang Kim}, \bibinfo{person}{Jaewoo
  Ahn}, {and} \bibinfo{person}{Gunhee Kim}.} \bibinfo{year}{2020}\natexlab{}.
\newblock \showarticletitle{Sequential Latent Knowledge Selection for
  Knowledge-Grounded Dialogue}.
\newblock \bibinfo{journal}{\emph{CoRR}}  \bibinfo{volume}{abs/2002.07510}
  (\bibinfo{year}{2020}).
\newblock
\showeprint[arxiv]{2002.07510}
\urldef\tempurl%
\url{https://arxiv.org/abs/2002.07510}
\showURL{%
\tempurl}


\bibitem[\protect\citeauthoryear{Kingma and Ba}{Kingma and Ba}{2015}]%
        {DBLP:journals/corr/KingmaB14}
\bibfield{author}{\bibinfo{person}{Diederik~P. Kingma} {and}
  \bibinfo{person}{Jimmy Ba}.} \bibinfo{year}{2015}\natexlab{}.
\newblock \showarticletitle{Adam: {A} Method for Stochastic Optimization}. In
  \bibinfo{booktitle}{\emph{3rd International Conference on Learning
  Representations, {ICLR} 2015, San Diego, CA, USA, May 7-9, 2015, Conference
  Track Proceedings}}.
\newblock
\urldef\tempurl%
\url{http://arxiv.org/abs/1412.6980}
\showURL{%
\tempurl}


\bibitem[\protect\citeauthoryear{Klein, Kim, Deng, Senellart, and Rush}{Klein
  et~al\mbox{.}}{2017}]%
        {klein-etal-2017-opennmt}
\bibfield{author}{\bibinfo{person}{Guillaume Klein}, \bibinfo{person}{Yoon
  Kim}, \bibinfo{person}{Yuntian Deng}, \bibinfo{person}{Jean Senellart}, {and}
  \bibinfo{person}{Alexander Rush}.} \bibinfo{year}{2017}\natexlab{}.
\newblock \showarticletitle{{O}pen{NMT}: Open-Source Toolkit for Neural Machine
  Translation}. In \bibinfo{booktitle}{\emph{Proceedings of {ACL} 2017, System
  Demonstrations}}. \bibinfo{publisher}{Association for Computational
  Linguistics}, \bibinfo{address}{Vancouver, Canada}, \bibinfo{pages}{67--72}.
\newblock
\urldef\tempurl%
\url{https://www.aclweb.org/anthology/P17-4012}
\showURL{%
\tempurl}


\bibitem[\protect\citeauthoryear{Li, Galley, Brockett, Gao, and Dolan}{Li
  et~al\mbox{.}}{2016}]%
        {DBLP:conf/naacl/LiGBGD16}
\bibfield{author}{\bibinfo{person}{Jiwei Li}, \bibinfo{person}{Michel Galley},
  \bibinfo{person}{Chris Brockett}, \bibinfo{person}{Jianfeng Gao}, {and}
  \bibinfo{person}{Bill Dolan}.} \bibinfo{year}{2016}\natexlab{}.
\newblock \showarticletitle{A Diversity-Promoting Objective Function for Neural
  Conversation Models}. In \bibinfo{booktitle}{\emph{{HLT-NAACL}}}.
  \bibinfo{publisher}{The Association for Computational Linguistics},
  \bibinfo{pages}{110--119}.
\newblock


\bibitem[\protect\citeauthoryear{Li, Niu, Meng, Feng, Li, and Zhou}{Li
  et~al\mbox{.}}{2019}]%
        {DBLP:conf/acl/LiNMFLZ19}
\bibfield{author}{\bibinfo{person}{Zekang Li}, \bibinfo{person}{Cheng Niu},
  \bibinfo{person}{Fandong Meng}, \bibinfo{person}{Yang Feng},
  \bibinfo{person}{Qian Li}, {and} \bibinfo{person}{Jie Zhou}.}
  \bibinfo{year}{2019}\natexlab{}.
\newblock \showarticletitle{Incremental Transformer with Deliberation Decoder
  for Document Grounded Conversations}. In \bibinfo{booktitle}{\emph{{ACL}
  {(1)}}}. \bibinfo{publisher}{Association for Computational Linguistics},
  \bibinfo{pages}{12--21}.
\newblock


\bibitem[\protect\citeauthoryear{Lin}{Lin}{2004}]%
        {lin2004rouge}
\bibfield{author}{\bibinfo{person}{Chin-Yew Lin}.}
  \bibinfo{year}{2004}\natexlab{}.
\newblock \showarticletitle{Rouge: A package for automatic evaluation of
  summaries}. In \bibinfo{booktitle}{\emph{Text summarization branches out}}.
  \bibinfo{pages}{74--81}.
\newblock


\bibitem[\protect\citeauthoryear{Liu, Chen, Ren, Feng, Liu, and Yin}{Liu
  et~al\mbox{.}}{2018}]%
        {DBLP:conf/acl/LiuFCRYL18}
\bibfield{author}{\bibinfo{person}{Shuman Liu}, \bibinfo{person}{Hongshen
  Chen}, \bibinfo{person}{Zhaochun Ren}, \bibinfo{person}{Yang Feng},
  \bibinfo{person}{Qun Liu}, {and} \bibinfo{person}{Dawei Yin}.}
  \bibinfo{year}{2018}\natexlab{}.
\newblock \showarticletitle{Knowledge Diffusion for Neural Dialogue
  Generation}. In \bibinfo{booktitle}{\emph{{ACL} {(1)}}}.
  \bibinfo{publisher}{Association for Computational Linguistics},
  \bibinfo{pages}{1489--1498}.
\newblock


\bibitem[\protect\citeauthoryear{Liu, Niu, Wu, and Wang}{Liu
  et~al\mbox{.}}{2019}]%
        {DBLP:journals/corr/abs-1903-10245}
\bibfield{author}{\bibinfo{person}{Zhibin Liu}, \bibinfo{person}{Zheng{-}Yu
  Niu}, \bibinfo{person}{Hua Wu}, {and} \bibinfo{person}{Haifeng Wang}.}
  \bibinfo{year}{2019}\natexlab{}.
\newblock \showarticletitle{Knowledge Aware Conversation Generation with
  Reasoning on Augmented Graph}.
\newblock \bibinfo{journal}{\emph{CoRR}}  \bibinfo{volume}{abs/1903.10245}
  (\bibinfo{year}{2019}).
\newblock


\bibitem[\protect\citeauthoryear{Luong, Pham, and Manning}{Luong
  et~al\mbox{.}}{2015}]%
        {DBLP:conf/emnlp/LuongPM15}
\bibfield{author}{\bibinfo{person}{Thang Luong}, \bibinfo{person}{Hieu Pham},
  {and} \bibinfo{person}{Christopher~D. Manning}.}
  \bibinfo{year}{2015}\natexlab{}.
\newblock \showarticletitle{Effective Approaches to Attention-based Neural
  Machine Translation}. In \bibinfo{booktitle}{\emph{Proceedings of the 2015
  Conference on Empirical Methods in Natural Language Processing, {EMNLP} 2015,
  Lisbon, Portugal, September 17-21, 2015}},
  \bibfield{editor}{\bibinfo{person}{Llu{\'{\i}}s M{\`{a}}rquez},
  \bibinfo{person}{Chris Callison{-}Burch}, \bibinfo{person}{Jian Su},
  \bibinfo{person}{Daniele Pighin}, {and} \bibinfo{person}{Yuval Marton}}
  (Eds.). \bibinfo{publisher}{The Association for Computational Linguistics},
  \bibinfo{pages}{1412--1421}.
\newblock
\urldef\tempurl%
\url{https://doi.org/10.18653/v1/d15-1166}
\showDOI{\tempurl}


\bibitem[\protect\citeauthoryear{Meng, Ren, Chen, Monz, Ma, and de~Rijke}{Meng
  et~al\mbox{.}}{2019}]%
        {DBLP:journals/corr/abs-1908-06449}
\bibfield{author}{\bibinfo{person}{Chuan Meng}, \bibinfo{person}{Pengjie Ren},
  \bibinfo{person}{Zhumin Chen}, \bibinfo{person}{Christof Monz},
  \bibinfo{person}{Jun Ma}, {and} \bibinfo{person}{Maarten de Rijke}.}
  \bibinfo{year}{2019}\natexlab{}.
\newblock \showarticletitle{RefNet: {A} Reference-aware Network for Background
  Based Conversation}.
\newblock \bibinfo{journal}{\emph{CoRR}}  \bibinfo{volume}{abs/1908.06449}
  (\bibinfo{year}{2019}).
\newblock


\bibitem[\protect\citeauthoryear{Mikolov, Sutskever, Chen, Corrado, and
  Dean}{Mikolov et~al\mbox{.}}{2013}]%
        {DBLP:conf/nips/MikolovSCCD13}
\bibfield{author}{\bibinfo{person}{Tomas Mikolov}, \bibinfo{person}{Ilya
  Sutskever}, \bibinfo{person}{Kai Chen}, \bibinfo{person}{Gregory~S. Corrado},
  {and} \bibinfo{person}{Jeffrey Dean}.} \bibinfo{year}{2013}\natexlab{}.
\newblock \showarticletitle{Distributed Representations of Words and Phrases
  and their Compositionality}. In \bibinfo{booktitle}{\emph{Advances in Neural
  Information Processing Systems 26: 27th Annual Conference on Neural
  Information Processing Systems 2013. Proceedings of a meeting held December
  5-8, 2013, Lake Tahoe, Nevada, United States}},
  \bibfield{editor}{\bibinfo{person}{Christopher J.~C. Burges},
  \bibinfo{person}{L{\'{e}}on Bottou}, \bibinfo{person}{Zoubin Ghahramani},
  {and} \bibinfo{person}{Kilian~Q. Weinberger}} (Eds.).
  \bibinfo{pages}{3111--3119}.
\newblock
\urldef\tempurl%
\url{http://papers.nips.cc/paper/5021-distributed-representations-of-words-and-phrases-and-their-compositionality}
\showURL{%
\tempurl}


\bibitem[\protect\citeauthoryear{Moghe, Arora, Banerjee, and Khapra}{Moghe
  et~al\mbox{.}}{2018}]%
        {DBLP:conf/emnlp/MogheABK18}
\bibfield{author}{\bibinfo{person}{Nikita Moghe}, \bibinfo{person}{Siddhartha
  Arora}, \bibinfo{person}{Suman Banerjee}, {and} \bibinfo{person}{Mitesh~M.
  Khapra}.} \bibinfo{year}{2018}\natexlab{}.
\newblock \showarticletitle{Towards Exploiting Background Knowledge for
  Building Conversation Systems}. In \bibinfo{booktitle}{\emph{{EMNLP}}}.
  \bibinfo{publisher}{Association for Computational Linguistics},
  \bibinfo{pages}{2322--2332}.
\newblock


\bibitem[\protect\citeauthoryear{Papineni, Roukos, Ward, and Zhu}{Papineni
  et~al\mbox{.}}{2002}]%
        {DBLP:conf/acl/PapineniRWZ02}
\bibfield{author}{\bibinfo{person}{Kishore Papineni}, \bibinfo{person}{Salim
  Roukos}, \bibinfo{person}{Todd Ward}, {and} \bibinfo{person}{Wei{-}Jing
  Zhu}.} \bibinfo{year}{2002}\natexlab{}.
\newblock \showarticletitle{Bleu: a Method for Automatic Evaluation of Machine
  Translation}. In \bibinfo{booktitle}{\emph{{ACL}}}.
  \bibinfo{publisher}{{ACL}}, \bibinfo{pages}{311--318}.
\newblock


\bibitem[\protect\citeauthoryear{Qin, Galley, Brockett, Liu, Gao, Dolan, Choi,
  and Gao}{Qin et~al\mbox{.}}{2019}]%
        {DBLP:conf/acl/QinGBLGDCG19}
\bibfield{author}{\bibinfo{person}{Lianhui Qin}, \bibinfo{person}{Michel
  Galley}, \bibinfo{person}{Chris Brockett}, \bibinfo{person}{Xiaodong Liu},
  \bibinfo{person}{Xiang Gao}, \bibinfo{person}{Bill Dolan},
  \bibinfo{person}{Yejin Choi}, {and} \bibinfo{person}{Jianfeng Gao}.}
  \bibinfo{year}{2019}\natexlab{}.
\newblock \showarticletitle{Conversing by Reading: Contentful Neural
  Conversation with On-demand Machine Reading}. In
  \bibinfo{booktitle}{\emph{{ACL} {(1)}}}. \bibinfo{publisher}{Association for
  Computational Linguistics}, \bibinfo{pages}{5427--5436}.
\newblock


\bibitem[\protect\citeauthoryear{Ren, Chen, Monz, Ma, and de~Rijke}{Ren
  et~al\mbox{.}}{2019}]%
        {DBLP:journals/corr/abs-1908-09528}
\bibfield{author}{\bibinfo{person}{Pengjie Ren}, \bibinfo{person}{Zhumin Chen},
  \bibinfo{person}{Christof Monz}, \bibinfo{person}{Jun Ma}, {and}
  \bibinfo{person}{Maarten de Rijke}.} \bibinfo{year}{2019}\natexlab{}.
\newblock \showarticletitle{Thinking Globally, Acting Locally: Distantly
  Supervised Global-to-Local Knowledge Selection for Background Based
  Conversation}.
\newblock \bibinfo{journal}{\emph{CoRR}}  \bibinfo{volume}{abs/1908.09528}
  (\bibinfo{year}{2019}).
\newblock


\bibitem[\protect\citeauthoryear{Sankar, Subramanian, Pal, Chandar, and
  Bengio}{Sankar et~al\mbox{.}}{2019}]%
        {DBLP:conf/acl/SankarSPCB19}
\bibfield{author}{\bibinfo{person}{Chinnadhurai Sankar},
  \bibinfo{person}{Sandeep Subramanian}, \bibinfo{person}{Chris Pal},
  \bibinfo{person}{Sarath Chandar}, {and} \bibinfo{person}{Yoshua Bengio}.}
  \bibinfo{year}{2019}\natexlab{}.
\newblock \showarticletitle{Do Neural Dialog Systems Use the Conversation
  History Effectively? An Empirical Study}. In
  \bibinfo{booktitle}{\emph{Proceedings of the 57th Conference of the
  Association for Computational Linguistics, {ACL} 2019, Florence, Italy, July
  28- August 2, 2019, Volume 1: Long Papers}}. \bibinfo{pages}{32--37}.
\newblock
\urldef\tempurl%
\url{https://www.aclweb.org/anthology/P19-1004/}
\showURL{%
\tempurl}


\bibitem[\protect\citeauthoryear{Seo, Kembhavi, Farhadi, and Hajishirzi}{Seo
  et~al\mbox{.}}{2017}]%
        {DBLP:conf/iclr/SeoKFH17}
\bibfield{author}{\bibinfo{person}{Min~Joon Seo}, \bibinfo{person}{Aniruddha
  Kembhavi}, \bibinfo{person}{Ali Farhadi}, {and} \bibinfo{person}{Hannaneh
  Hajishirzi}.} \bibinfo{year}{2017}\natexlab{}.
\newblock \showarticletitle{Bidirectional Attention Flow for Machine
  Comprehension}. In \bibinfo{booktitle}{\emph{{ICLR} (Poster)}}.
  \bibinfo{publisher}{OpenReview.net}.
\newblock


\bibitem[\protect\citeauthoryear{Serban, Sordoni, Bengio, Courville, and
  Pineau}{Serban et~al\mbox{.}}{2016}]%
        {DBLP:conf/aaai/SerbanSBCP16}
\bibfield{author}{\bibinfo{person}{Iulian~Vlad Serban},
  \bibinfo{person}{Alessandro Sordoni}, \bibinfo{person}{Yoshua Bengio},
  \bibinfo{person}{Aaron~C. Courville}, {and} \bibinfo{person}{Joelle Pineau}.}
  \bibinfo{year}{2016}\natexlab{}.
\newblock \showarticletitle{Building End-To-End Dialogue Systems Using
  Generative Hierarchical Neural Network Models}. In
  \bibinfo{booktitle}{\emph{Proceedings of the Thirtieth {AAAI} Conference on
  Artificial Intelligence, February 12-17, 2016, Phoenix, Arizona, {USA}}}.
  \bibinfo{pages}{3776--3784}.
\newblock
\urldef\tempurl%
\url{http://www.aaai.org/ocs/index.php/AAAI/AAAI16/paper/view/11957}
\showURL{%
\tempurl}


\bibitem[\protect\citeauthoryear{Serban, Sordoni, Lowe, Charlin, Pineau,
  Courville, and Bengio}{Serban et~al\mbox{.}}{2017}]%
        {DBLP:conf/aaai/SerbanSLCPCB17}
\bibfield{author}{\bibinfo{person}{Iulian~Vlad Serban},
  \bibinfo{person}{Alessandro Sordoni}, \bibinfo{person}{Ryan Lowe},
  \bibinfo{person}{Laurent Charlin}, \bibinfo{person}{Joelle Pineau},
  \bibinfo{person}{Aaron~C. Courville}, {and} \bibinfo{person}{Yoshua Bengio}.}
  \bibinfo{year}{2017}\natexlab{}.
\newblock \showarticletitle{A Hierarchical Latent Variable Encoder-Decoder
  Model for Generating Dialogues}. In \bibinfo{booktitle}{\emph{Proceedings of
  the Thirty-First {AAAI} Conference on Artificial Intelligence, February 4-9,
  2017, San Francisco, California, {USA}}}. \bibinfo{pages}{3295--3301}.
\newblock
\urldef\tempurl%
\url{http://aaai.org/ocs/index.php/AAAI/AAAI17/paper/view/14567}
\showURL{%
\tempurl}


\bibitem[\protect\citeauthoryear{Sutskever, Vinyals, and Le}{Sutskever
  et~al\mbox{.}}{2014}]%
        {DBLP:conf/nips/SutskeverVL14}
\bibfield{author}{\bibinfo{person}{Ilya Sutskever}, \bibinfo{person}{Oriol
  Vinyals}, {and} \bibinfo{person}{Quoc~V. Le}.}
  \bibinfo{year}{2014}\natexlab{}.
\newblock \showarticletitle{Sequence to Sequence Learning with Neural
  Networks}. In \bibinfo{booktitle}{\emph{{NIPS}}}.
  \bibinfo{pages}{3104--3112}.
\newblock


\bibitem[\protect\citeauthoryear{Tian, Bi, Lee, Xue, Song, Liu, and Zhang}{Tian
  et~al\mbox{.}}{2020}]%
        {DBLP:journals/corr/abs-2005-06128}
\bibfield{author}{\bibinfo{person}{Zhiliang Tian}, \bibinfo{person}{Wei Bi},
  \bibinfo{person}{Dongkyu Lee}, \bibinfo{person}{Lanqing Xue},
  \bibinfo{person}{Yiping Song}, \bibinfo{person}{Xiaojiang Liu}, {and}
  \bibinfo{person}{Nevin~L. Zhang}.} \bibinfo{year}{2020}\natexlab{}.
\newblock \showarticletitle{Response-Anticipated Memory for On-Demand Knowledge
  Integration in Response Generation}.
\newblock \bibinfo{journal}{\emph{CoRR}}  \bibinfo{volume}{abs/2005.06128}
  (\bibinfo{year}{2020}).
\newblock
\showeprint[arxiv]{2005.06128}
\urldef\tempurl%
\url{https://arxiv.org/abs/2005.06128}
\showURL{%
\tempurl}


\bibitem[\protect\citeauthoryear{Vaswani, Shazeer, Parmar, Uszkoreit, Jones,
  Gomez, Kaiser, and Polosukhin}{Vaswani et~al\mbox{.}}{2017}]%
        {DBLP:conf/nips/VaswaniSPUJGKP17}
\bibfield{author}{\bibinfo{person}{Ashish Vaswani}, \bibinfo{person}{Noam
  Shazeer}, \bibinfo{person}{Niki Parmar}, \bibinfo{person}{Jakob Uszkoreit},
  \bibinfo{person}{Llion Jones}, \bibinfo{person}{Aidan~N. Gomez},
  \bibinfo{person}{Lukasz Kaiser}, {and} \bibinfo{person}{Illia Polosukhin}.}
  \bibinfo{year}{2017}\natexlab{}.
\newblock \showarticletitle{Attention is All you Need}. In
  \bibinfo{booktitle}{\emph{{NIPS}}}. \bibinfo{pages}{5998--6008}.
\newblock


\bibitem[\protect\citeauthoryear{Xia, Tian, Wu, Lin, Qin, Yu, and Liu}{Xia
  et~al\mbox{.}}{2017}]%
        {DBLP:conf/nips/XiaTWLQYL17}
\bibfield{author}{\bibinfo{person}{Yingce Xia}, \bibinfo{person}{Fei Tian},
  \bibinfo{person}{Lijun Wu}, \bibinfo{person}{Jianxin Lin},
  \bibinfo{person}{Tao Qin}, \bibinfo{person}{Nenghai Yu}, {and}
  \bibinfo{person}{Tie{-}Yan Liu}.} \bibinfo{year}{2017}\natexlab{}.
\newblock \showarticletitle{Deliberation Networks: Sequence Generation Beyond
  One-Pass Decoding}. In \bibinfo{booktitle}{\emph{{NIPS}}}.
  \bibinfo{pages}{1784--1794}.
\newblock


\bibitem[\protect\citeauthoryear{Ye, Lo, Su, and Chen}{Ye
  et~al\mbox{.}}{2019}]%
        {DBLP:journals/corr/abs-1903-09813}
\bibfield{author}{\bibinfo{person}{Hao{-}Tong Ye}, \bibinfo{person}{Kai{-}Ling
  Lo}, \bibinfo{person}{Shang{-}Yu Su}, {and} \bibinfo{person}{Yun{-}Nung
  Chen}.} \bibinfo{year}{2019}\natexlab{}.
\newblock \showarticletitle{Knowledge-Grounded Response Generation with Deep
  Attentional Latent-Variable Model}.
\newblock \bibinfo{journal}{\emph{CoRR}}  \bibinfo{volume}{abs/1903.09813}
  (\bibinfo{year}{2019}).
\newblock


\bibitem[\protect\citeauthoryear{Young, Cambria, Chaturvedi, Zhou, Biswas, and
  Huang}{Young et~al\mbox{.}}{2018}]%
        {DBLP:conf/aaai/YoungCCZBH18}
\bibfield{author}{\bibinfo{person}{Tom Young}, \bibinfo{person}{Erik Cambria},
  \bibinfo{person}{Iti Chaturvedi}, \bibinfo{person}{Hao Zhou},
  \bibinfo{person}{Subham Biswas}, {and} \bibinfo{person}{Minlie Huang}.}
  \bibinfo{year}{2018}\natexlab{}.
\newblock \showarticletitle{Augmenting End-to-End Dialogue Systems With
  Commonsense Knowledge}. In \bibinfo{booktitle}{\emph{{AAAI}}}.
  \bibinfo{publisher}{{AAAI} Press}, \bibinfo{pages}{4970--4977}.
\newblock


\bibitem[\protect\citeauthoryear{Zhang, Ren, and de~Rijke}{Zhang
  et~al\mbox{.}}{2019}]%
        {DBLP:journals/corr/abs-1906-06685}
\bibfield{author}{\bibinfo{person}{Yangjun Zhang}, \bibinfo{person}{Pengjie
  Ren}, {and} \bibinfo{person}{Maarten de Rijke}.}
  \bibinfo{year}{2019}\natexlab{}.
\newblock \showarticletitle{Improving Background Based Conversation with
  Context-aware Knowledge Pre-selection}.
\newblock \bibinfo{journal}{\emph{CoRR}}  \bibinfo{volume}{abs/1906.06685}
  (\bibinfo{year}{2019}).
\newblock


\bibitem[\protect\citeauthoryear{Zhao, Zhao, and Esk{\'{e}}nazi}{Zhao
  et~al\mbox{.}}{2017}]%
        {DBLP:conf/acl/ZhaoZE17}
\bibfield{author}{\bibinfo{person}{Tiancheng Zhao}, \bibinfo{person}{Ran Zhao},
  {and} \bibinfo{person}{Maxine Esk{\'{e}}nazi}.}
  \bibinfo{year}{2017}\natexlab{}.
\newblock \showarticletitle{Learning Discourse-level Diversity for Neural
  Dialog Models using Conditional Variational Autoencoders}. In
  \bibinfo{booktitle}{\emph{{ACL} {(1)}}}. \bibinfo{publisher}{Association for
  Computational Linguistics}, \bibinfo{pages}{654--664}.
\newblock


\bibitem[\protect\citeauthoryear{Zhao, Tao, Wu, Xu, Zhao, and Yan}{Zhao
  et~al\mbox{.}}{2019}]%
        {DBLP:conf/ijcai/ZhaoTWX0Y19}
\bibfield{author}{\bibinfo{person}{Xueliang Zhao}, \bibinfo{person}{Chongyang
  Tao}, \bibinfo{person}{Wei Wu}, \bibinfo{person}{Can Xu},
  \bibinfo{person}{Dongyan Zhao}, {and} \bibinfo{person}{Rui Yan}.}
  \bibinfo{year}{2019}\natexlab{}.
\newblock \showarticletitle{A Document-grounded Matching Network for Response
  Selection in Retrieval-based Chatbots}. In
  \bibinfo{booktitle}{\emph{{IJCAI}}}. \bibinfo{publisher}{ijcai.org},
  \bibinfo{pages}{5443--5449}.
\newblock


\bibitem[\protect\citeauthoryear{Zheng and Zhou}{Zheng and Zhou}{2019}]%
        {DBLP:conf/cikm/ZhengZ19}
\bibfield{author}{\bibinfo{person}{Wen Zheng} {and} \bibinfo{person}{Ke Zhou}.}
  \bibinfo{year}{2019}\natexlab{}.
\newblock \showarticletitle{Enhancing Conversational Dialogue Models with
  Grounded Knowledge}. In \bibinfo{booktitle}{\emph{{CIKM}}}.
  \bibinfo{publisher}{{ACM}}, \bibinfo{pages}{709--718}.
\newblock


\bibitem[\protect\citeauthoryear{Zhou, Young, Huang, Zhao, Xu, and Zhu}{Zhou
  et~al\mbox{.}}{2018b}]%
        {DBLP:conf/ijcai/ZhouYHZXZ18}
\bibfield{author}{\bibinfo{person}{Hao Zhou}, \bibinfo{person}{Tom Young},
  \bibinfo{person}{Minlie Huang}, \bibinfo{person}{Haizhou Zhao},
  \bibinfo{person}{Jingfang Xu}, {and} \bibinfo{person}{Xiaoyan Zhu}.}
  \bibinfo{year}{2018}\natexlab{b}.
\newblock \showarticletitle{Commonsense Knowledge Aware Conversation Generation
  with Graph Attention}. In \bibinfo{booktitle}{\emph{{IJCAI}}}.
  \bibinfo{publisher}{ijcai.org}, \bibinfo{pages}{4623--4629}.
\newblock


\bibitem[\protect\citeauthoryear{Zhou, Prabhumoye, and Black}{Zhou
  et~al\mbox{.}}{2018a}]%
        {DBLP:conf/emnlp/ZhouPB18}
\bibfield{author}{\bibinfo{person}{Kangyan Zhou}, \bibinfo{person}{Shrimai
  Prabhumoye}, {and} \bibinfo{person}{Alan~W. Black}.}
  \bibinfo{year}{2018}\natexlab{a}.
\newblock \showarticletitle{A Dataset for Document Grounded Conversations}. In
  \bibinfo{booktitle}{\emph{{EMNLP}}}. \bibinfo{publisher}{Association for
  Computational Linguistics}, \bibinfo{pages}{708--713}.
\newblock


\end{thebibliography}

\end{document}